\documentclass[conference,anonymous=false]{IEEEtran}
\IEEEoverridecommandlockouts
\usepackage{algorithm}
\usepackage{algorithmic}
\usepackage{amsmath,amssymb,amsfonts}
\usepackage{array}
\usepackage{cite}
\usepackage{graphicx}
\usepackage[hyphens]{url}
\usepackage{hyperref}
\usepackage[hyphenbreaks]{breakurl}
\usepackage{listings}
\usepackage{multirow,tabularx}
\usepackage{stfloats}
\usepackage[caption=false,font=normalsize,labelfont=sf,textfont=sf]{subfig}
\usepackage{textcomp}
\usepackage{verbatim}
\usepackage{xcolor}
\def\BibTeX{{\rm B\kern-.05em{\sc i\kern-.025em b}\kern-.08em
    T\kern-.1667em\lower.7ex\hbox{E}\kern-.125emX}}
\lstdefinelanguage{Toml}{
  comment=[l]{\#},
  keywords={true, false},
  morestring=[b]{"},
}
\hypersetup{
  colorlinks=true,
}
\begin{document}

\title{Recording and Describing Poker Hands}

% \author{\textit{Redacted for Double-Blind Review}}
\author{\IEEEauthorblockN{Juho Kim}
\IEEEauthorblockA{\textit{Faculty of Applied Science and Engineering} \\
\textit{University of Toronto}\\
Toronto, Ontario, Canada \\
\href{mailto:juho.kim@mail.utoronto.ca}{juho.kim@mail.utoronto.ca}}
}

\IEEEoverridecommandlockouts
\IEEEpubid{\makebox[\columnwidth]{ 979-8-3503-5067-8/24/\$31.00~\copyright2024 IEEE \hfill} %–> insert the copyright option applicable from above.
\hspace{\columnsep}\makebox[\columnwidth]{ }}

\maketitle

% Add the following code after the \\make title command:
\IEEEpubidadjcol

\begin{abstract}
This paper introduces the Poker Hand History (PHH) file format, designed to standardize the recording of poker hands across different game variants. Despite poker's widespread popularity in the mainstream culture as a mind sport and its prominence in the field of artificial intelligence (AI) research as a benchmark for imperfect information AI agents, it lacks a consistent format that humans can use to document poker hands across different variants that can also easily be parsed by machines. To address this gap in the literature, we propose the PHH format which provides a concise human-readable machine-friendly representation of hand history that comprehensively captures various details of the hand, ranging from initial game parameters and actions to contextual parameters including but not limited to the venue, players, and time control information. In the supplementary, we provide 10,088 hands covering 11 different variants in the PHH format. The full specification is available on \url{https://github.com/uoftcprg/phh-std}
\end{abstract}


\section{Introduction}

Poker is a popular card game with a stochastic nature and imperfect information aspect that has not only captured mainstream attention as a leading mind sport but also served as a benchmark of artificial intelligence (AI) research, namely in the development of imperfect information AI agents.

Pioneering breakthroughs in imperfect information game-solving algorithms have often been accompanied by their applications in state-of-the-art poker AI agents as seen in the example of Rhode Island hold 'em \cite{DBLP:conf/aaai/GilpinS05}, fixed-limit Texas hold 'em \cite{DBLP:journals/cacm/BowlingBJT17}, heads-up (2-player) no-limit Texas hold 'em \cite{doi:10.1126/science.aam6960, doi:10.1126/science.aao1733}, and 6-max (6-player) no-limit Texas hold 'em \cite{doi:10.1126/science.aay2400}. Successive state-of-the-art algorithms were shown to solve games of progressively larger game tree sizes.

While impressive, the applications of these algorithms were mostly limited to the Texas hold 'em variant which ignores the vast diversity of poker. Indeed, there exists countless variants of poker that introduce different rules and actions not present in Texas hold 'em \cite{2023wsoptourneys, 2023wsopliveaction} that can serve as new challenges in the development of imperfect information AI agents. For instance, Omaha variants involve four hole cards per player instead of simply two, which increases the range of possible holdings of opponents, further increasing the game tree size. Another popular variant, lowball draw, introduces shifting hidden information depending on the discards and draws unlike Texas hold 'em where the players' hole cards remain unchanged throughout the hand.

While it is true that Texas hold 'em is the most prominent variant of poker, as it is the variant of choice in the World Series of Poker (WSOP) Main Event, the largest poker tournament in the world, there is a substantial interest in the market and the industry on other poker variants. For instance, of the 94 in-person bracelet events in the 2023 WSOP, only around 51 events (54.26\%) play no-limit Texas hold 'em exclusively. \cite{2023wsopschedule} This diversity of the game differs from other popular board games like chess, shogi, or go which has a universally accepted ``standard'' variant.

In addition, various initial state assumptions made by the state-of-the-art poker AI agents make it completely inapplicable to real-life gameplays where the assumptions are rarely true. All the recent breakthroughs in poker AI agents deal with uniform starting stack sizes, where every player has an identical number of chips at the beginning of the game, in addition to just one configuration of antes and forced bets like blinds or straddles. Not to mention, their solutions are only meant to be used in either heads-up or 6-max settings. These assumptions, of course, are unrealistic, and taking these variations into account vastly increases the game tree size. It is therefore uncertain if existing techniques for solving imperfect information games can handle game trees of such size. The implications of the existence of such an algorithm would be immense, given that the vast number of real-world problems such as stock markets, auctions, and diplomacy are examples of imperfect information games like poker.

The lack of exploration in the aforementioned areas of computer poker can partly be explained by the lack of reliable open-source multivariant poker tooling. Our recent work on PokerKit \cite{10287546} aimed to provide a robust implementation of poker simulation and hand evaluation suite to address this absence and has since established itself as a popular tool in computational poker.

However, our previous efforts in the digitalization of poker were only partially complete, as we provided no means to record poker hands for persistent storage and subsequent analysis, a crucial component in the training and evaluation of AI agents. This is partly because there does not exist a standardized format for recording poker games that is not only human-friendly but also easily parsable by computer software akin to the portable game notation (PGN) for chess \cite{pgn} or smart game format (SGF) for go \cite{sgf}.

This paper proposes a novel open-source Poker Hand History (PHH) file format designed to provide a consistent method to annotate poker hands across different variants in addition to capturing other details of the game including but not limited to the venue, players, and time control information that can aid in domains such as poker AI agent development, historical analysis, and data referencing.

In the next section, we provide a motivational example that showcase the versatility of the PHH notation. This is followed by an exploration of the related works in the literature that introduce alternatives to our proposal with which we later run benchmarks for comparison purposes. Then, we provide the detailed specification of the PHH format, listing the required and optional fields used to comprehensively describe poker hands in addition to various recommendations in file formatting, style guides, and parser implementation.

In the supplementary, 10,088 sample poker hands, covering 11 different poker variants and composed of small selections of historical or notable hands, all 83 televised hands in the 2023 WSOP Event \#43: \$50,000 Poker Players Championship (PPC) | Day 5 \cite{2023wsopevent43}, and the hands played by Pluribus in Brown and Sandholm \cite{doi:10.1126/science.aay2400}, are provided in the PHH file format.

\section{Motivational Example}

\begin{figure}[htbp]
  \centering
  \begin{lstlisting}[language=Toml]
# A bad beat between Yockey and Arieh.
variant = "F2L3D"
antes = [0, 0, 0, 0]
blinds_or_straddles = [
  75000, 150000, 0, 0,
]
small_bet = 150000
big_bet = 300000
starting_stacks = [
  1180000, 4340000, 5910000, 10765000,
]
actions = [
  "d dh p1 7h6c4c3d2c",  # Yockey
  "d dh p2 ??????????",  # Hui
  "d dh p3 ??????????",  # Esposito
  "d dh p4 AsQs6s5c3c",  # Arieh
  "p3 f",  # Esposito
  "p4 cbr 300000",  # Arieh
  "p1 cbr 450000",  # Yockey
  "p2 f",  # Hui
  "p4 cc",  # Arieh
  "p1 sd",  # First draw; Yockey
  "p4 sd AsQs",  # Arieh
  "d dh p4 2hQh",  # Arieh
  "p1 cbr 150000",  # Yockey
  "p4 cc",  # Arieh
  "p1 sd",  # Second draw; Yockey
  "p4 sd Qh",  # Arieh
  "d dh p4 4d",  # Arieh
  "p1 cbr 300000",  # Yockey
  "p4 cc",  # Arieh
  "p1 sd",  # Third draw; Yockey
  "p4 sd 6s",  # Arieh
  "d dh p4 7c",  # Arieh
  "p1 cbr 280000",  # Yockey
  "p4 cc",  # Arieh
  "p1 sm 7h6c4c3d2c",  # Showdown; Yockey
  "p4 sm 2h4d7c5c3c",  # Arieh
]
event = "2019 WSOP Event #58"
city = "Las Vegas"
region = "Nevada"
country = "United States of America"
day = 28
month = 6
year = 2019
players = [
  "Bryce Yockey", "Phil Hui",
  "John Esposito", "Josh Arieh",
]
  \end{lstlisting}
  \caption{A bad beat between Josh Arieh and Bryce Yockey \cite{ariehyockey2019} in the PHH file format.}
  \label{fig:ariehyockey2019}
\end{figure}

Figure \ref{fig:ariehyockey2019} shows an example of a poker hand digitalized into the PHH format from a bad beat from the final table of the 2019 WSOP Event \#52: \$50,000 PPC on Day 5 of the event between Josh Arieh and Bryce Yockey \cite{ariehyockey2019}.

The PHH format aims to be both human-readable and writable while maintaining easy parsability by relevant software systems. In this format, the hands are described through the definitions of various key/values, along with optional comments throughout the file.

\section{Related Works}

In poker literature, it has historically been the norm to describe poker games verbally as exemplified in the news articles covering the aforementioned hands \cite{dwanivey2009, ariehyockey2019}. Such a practice, while fit for human readership, presents significant challenges for parsing and computational analysis. While the description could be rich with information like the names of players involved, the backstory of the hand, et cetera, it could also lack nuances and details that may be of interest to the readers such as the exact stack sizes, bet sizes, and mucked hand values. The subjective formatting may introduce bias and inconsistencies unsuitable for scientific use cases.

One of the earliest attempts to introduce a form of structure to hand summaries was the Mike Caro University Poker Chart proposed by Mike Caro \cite{caro2003}. His tabular format where players and betting rounds are represented as a table adds a form of structure to the data that makes it easier for humans to follow. In addition, various mathematical symbols and operations such as ``=,'' ``$\sqrt{}$,'' ``--,'' exponentiation, and more were used to concisely represent each betting action. Caro's format serves as an innovative breakthrough when it comes to improvements in visual clarity and consistency. Unfortunately, various descriptions of poker such as stakes remain verbal, and are designed primarily for human-to-human communication. In hindsight, it failed to become a standard in the poker community, especially in the age of computer poker where machine interpretability is another vital angle that must be considered when designing such standards.

PokerStars, one of the most popular online poker platforms, saves a log of the game history in their proprietary PokerStars Hand History format \cite{pokerstarshandreplay}. While the logs are designed for human readership and record-keeping, they are in plaintext format and are highly structured and consistent, allowing various poker analytic tools in the industry such as Holdem Manager, Poker Copilot, and PokerSnowie to parse and analyze a player's hand history reliably. The logged details include granular information down to the amounts of antes and blinds being posted by each player. This fine detail unfortunately makes it too verbose to be hand-typed by a human while being too specialized for use in video poker, namely PokerStars. For instance, some digital archiving use cases may require additional or complete descriptions of the game like the final stacks, time control, et cetera with features of the game not available on PokerStars. Not only that, the verbal nature of the format introduces unnecessary verbosity that wastes storage space. This flaw is further exacerbated by its lack of sufficient specification or documentation which introduces substantial difficulties when developing datasets in the PokerStars format. This is evident by the fact that there has not been any notable community effort to digitalize any historical poker hands.

The dataset of hands played by Pluribus, in the supplementary of Brown and Sandholm \cite{doi:10.1126/science.aay2400}, is formatted extremely concisely where a single hand is summarized into a dense line of characters. This format includes the bare minimum description of actions to comprehensively represent a no-limit Texas hold 'em hand assuming uniform starting stacks and standard double blinds. However, since each action notation does not specify the actor, the hands are difficult to follow by a human reader. In addition, the format is limited to only a single variant: no-limit Texas hold 'em, and makes critical assumptions such as the identical and fixed starting stack sizes of 100 big blinds and the number of players being 6. On top of this, it lacks options to include various metadata outside of the game state descriptions that may be of interest for data referencing purposes.

Standardized recording formats of other prominent board games such as PGN format for chess \cite{pgn} and SGF for go \cite{sgf} provide a good direction that can be followed for the game of poker. These formats specify tags or properties that can be omitted depending on the needs of the end user for encoding information not limited to the game state or the list of actions but also numerous extra information designed to give extra context to the game played like the location or the commentary just to name a few. Our PHH format takes inspiration from the aforementioned approaches and adopts them for use in the recording of poker hand history.

\section{Objective}

An ideal representation of poker hands should not only be concise but also easily readable and writable by human readers. It should also support only describing the bare minimum information necessary to comprehensively describe a hand while facilitating options to give rich details for various use cases beyond simple action tracking. For convenience in parser implementations, the formatting must also be simply structured in a consistent manner. In addition, the format must be powerful enough to describe a wide variety of poker games played in formal settings. Such selections should cover games played in reputable poker tournaments around the world while the inclusion of various eccentric or esoteric variants is not as necessary.

\section{Specification}

The PHH format is a derivative of the Tom's Obvious, Minimal Language (TOML) format \cite{toml}. This design allows PHH files to take advantage of TOML format's type system and be human-readable and writable while maintaining easy parsability by software systems. This differs from historical game file formats like PGN \cite{pgn} or SGF \cite{sgf} which use in-house formatting.

The PHH format puts restrictions on the naming and types of the key/values, with which the poker games are described. These definitions, named ``fields,'' fall under two classifications. The first involves state construction and progression and must be specified. The second describes various miscellaneous information about the game and may be omitted by the annotator.

\subsection{Error Handling}

In the implementation of the PHH parser, the program should report an error when any violation of the specification is encountered. The parser may stop parsing the file, ignore the problematic field or action, or proceed as normal.

\subsection{Position}

The fields that have a value of type array where each element corresponds to a particular player must have their elements ordered identically to the ordering of the players and have lengths equal to the number of players. The ordering of the players is table agnostic, that is, it is purely based on position.

The first player in the PHH format is typically the very first person being dealt the hole card by the dealer while the last player in the PHH format is typically the person being dealt the last hole card. This is assuming a proper dealing order is followed. In general, the ordering of the players must represent a clockwise order.

In typical button games such as no-limit Texas hold 'em or pot-limit Omaha hold 'em, the first player must be in the small blind (or big blind if heads-up) while the last player must be in position (i.e. has the button). In non-button games like stud, there is no strict notion of position. In such games, the first player should be the one in the dealer's immediate left while the last player should be in the dealer's immediate right.

\subsection{Rules}

As poker is a game with a diverse set of rules that often differ based on regions or events, it is impossible to point to anything as an authoritative documentation of poker rules that must be followed both by the annotator and the parser. The parser implementation should be designed to handle actions that may be potential transgressions in certain poker rule sets, either silently or with a warning.

While annotating the fields, especially the ``actions'' field, the author should follow the specific set of rules followed by the game being annotated. Where applicable or possible, the author is encouraged to conform to the latest WSOP Tournament Rules and WSOP Live Action Rules of which the most recently released versions as of the writing of this paper are for the 2023 WSOP \cite{2023wsoptourneys, 2023wsopliveaction}. If there is any contradiction between the two documents, the tournament rules take precedence, as they impose stricter guidelines on poker playing. When there is an ambiguity in both documents, it is to be considered as an undefined behavior. In such a case, the annotator should adhere to a historically informed custom of poker playing.

\section{Required Fields}

\begin{table}[!ht]
  \caption{The 11 variant codes and the corresponding full variant names. \label{tab:variants}}
  \centering
  \begin{tabular}{|c|c|}
    \hline
    Code & Name \\
    \hline\hline
    ``FT'' & Fixed-limit Texas hold 'em \\
    \hline
    ``NT'' & No-limit Texas hold 'em \\
    \hline
    ``NS'' & No-limit short-deck hold 'em \\
    \hline
    ``PO'' & Pot-limit Omaha hold 'em \\
    \hline
    ``FO/8'' & Fixed-limit Omaha hold 'em high/low-split eight or better \\
    \hline
    ``F7S'' & Fixed-limit seven card stud \\
    \hline
    ``F7S/8'' & Fixed-limit seven card stud high/low-split eight or better \\
    \hline
    ``FR'' & Fixed-limit razz \\
    \hline
    ``N2L1D'' & No-limit deuce-to-seven lowball single draw \\
    \hline
    ``F2L3D'' & Fixed-limit deuce-to-seven lowball triple draw \\
    \hline
    ``FB'' & Fixed-limit badugi \\
    \hline
  \end{tabular}
\end{table}

\begin{table*}[!ht]
  \caption{The composition of required fields for each variant. \label{tab:requiredfields}}
  \centering
  \small
  \begin{tabular}{|c||c|c|c|c|c|c|c|c|}
    \hline
    Variant & Antes & Blinds or Straddles & Bring-in & Small bet & Big bet & Min bet & Starting stacks & Actions \\
    \hline\hline
    ``FT'' & yes & yes & no & yes & yes & no & yes & yes \\
    \hline
    ``NT'' & yes & yes & no & no & no & yes & yes & yes \\
    \hline
    ``NS'' & yes & yes & no & no & no & yes & yes & yes \\
    \hline
    ``PO'' & yes & yes & no & no & no & yes & yes & yes \\
    \hline
    ``FO/8'' & yes & yes & no & yes & yes & no & yes & yes \\
    \hline
    ``F7S'' & yes & no & yes & yes & yes & no & yes & yes \\
    \hline
    ``F7S/8'' & yes & no & yes & yes & yes & no & yes & yes \\
    \hline
    ``FR'' & yes & no & yes & yes & yes & no & yes & yes \\
    \hline
    ``N2L1D'' & yes & yes & no & no & no & yes & yes & yes \\
    \hline
    ``F2L3D'' & yes & yes & no & yes & yes & no & yes & yes \\
    \hline
    ``FB'' & yes & yes & no & yes & yes & no & yes & yes \\
    \hline
  \end{tabular}
\end{table*}

\begin{table}[!ht]
  \caption{The required fields and their TOML native types. \label{tab:requiredfieldtypes}}
  \centering
  \begin{tabular}{|c|c|c|}
    \hline
    Field & Name & TOML Native Type \\
    \hline\hline
    Variant code & variant & String \\
    \hline
    Antes & antes & Array of integers or floats \\
    \hline
    Blinds/Straddles & blinds\_or\_straddles & Array of integers or floats \\
    \hline
    Bring-in & bring\_in & Integer or float \\
    \hline
    Small bet & small\_bet & Integer or float \\
    \hline
    Big bet & big\_bet & Integer or float \\
    \hline
    Min bet & min\_bet & Integer or float \\
    \hline
    Starting stacks & starting\_stacks & Array of integers, floats, or null \\
    \hline
    Actions & actions & Array of strings \\
    \hline
  \end{tabular}
\end{table}

\begin{figure}[htbp]
  \centering
  \begin{lstlisting}[language=Toml]
# Zero antes.
antes = [0, 0, 0, 0, 0]

# Antes of 1.
antes = [1, 1]

# Big blind ante (heads-up)
antes = [0.0, 3.0]

# Big blind ante
antes = [0.0, 3.0, 0.0]

# Button ante
antes = [0, 0, 0, 10]

# Small and big blind (heads-up)
blinds_or_straddles = [1, 2]

# Small and big blind
blinds_or_straddles = [1, 2, 0, 0, 0, 0]

# Under-the-gun (UTG) straddle
blinds_or_straddles = [1, 2, 4, 0]

# UTG and UTG+1 straddle
blinds_or_straddles = [1, 2, 4, 8, 0, 0]

# Button straddle
blinds_or_straddles = [1, 2, 0, 0, 0, 4]
  \end{lstlisting}
  \caption{Example definitions in PHH format.}
  \label{fig:definitions}
\end{figure}

The required fields represent a handful of fields that the annotator must define to digitalize a poker hand and therefore form a valid PHH file. The required fields are either used during the initial state construction or the subsequent state progression.

In this paper, 11 variants are explored, each of which is written in an abbreviated form, as tabulated in Table \ref{tab:variants}. These variants include all 9 played in the 2023 WSOP Event 43: \$50,000 PPC \cite{2023wsopevent43}, in addition to 2 others like short-deck and badugi in which the poker community has shown growing interest. These were chosen as they adequately represent the diverse nature of poker variants and demonstrate the flexibility of the specification laid out in this paper which is powerful enough to accommodate all variants in the 2023 WSOP Tournament Rules \cite{2023wsoptourneys}. Those not mentioned in Table \ref{tab:variants} can simply be integrated by introducing new variant codes to be associated with, and they will be introduced on a need-by-need basis based on community feedback in the future version of the specification.

The actual composition of the required fields varies, depending on the played variant. The required field statuses for each variant are described in Table \ref{tab:requiredfields}, and the fields' native TOML types are specified in Table \ref{tab:requiredfieldtypes}.

The fields ``antes'' and ``blinds\_or\_straddles'' represent the antes and forced bets (except the bring-in), respectively, posted by each of the players in the hand. Note that antes and blinds are reversed in heads-up games by rule. Example definitions are shown in Figure \ref{fig:definitions}. The ``blinds\_or\_straddles'' are forced bets only relevant in button games while the ``bring\_in'' is a type of forced bet only used in stud game variants.

The ``small\_bet'' and ``min\_bet'' fields denote the minimum bets in the first and last few betting rounds in fixed-limit games. In no-limit or pot-limit games, these two values are, by rule, identical and therefore combined into a single ``min\_bet'' field.

The ``starting\_stacks'' field denotes the starting stacks of each player and is an array of positive integers or floats of length equal to the number of players. All stack sizes must strictly be non-zero. Unknown stack values can be denoted as ``null''.

\begin{table}[!ht]
  \caption{Action notations in PHH File Format. \label{tab:grammar}}
  \centering
  \begin{tabular}{|c|c|c|}
    \hline
    Action & Grammar \\
    \hline\hline
    Dealing community cards & \text{d db card(s)}[\text{ \# Commentary}] \\
    \hline
    Dealing down/up cards & \text{d dh p}\textit{n}\text{ card(s)}[\text{ \# Commentary}] \\
    \hline
    Bringing in & \text{p}\textit{n}\text{ pb}[\text{ \# Commentary}] \\
    \hline
    Completing/Betting/Raising to & \text{p}\textit{n}\text{ cbr amount}[\text{ \# Commentary}] \\
    \hline
    Checking/Calling & \text{p}\textit{n}\text{ cc}[\text{ \# Commentary}] \\
    \hline
    Folding & \text{p}\textit{n}\text{ f}[\text{ \# Commentary}] \\
    \hline
    Standing pat/Discarding & \text{p}\textit{n}\text{ sd}[\text{ card(s)}][\text{ \# Commentary}] \\
    \hline
    Showing/Mucking their hole cards & \text{p}\textit{n}\text{ sm}[\text{ card(s)}][\text{ \# Commentary}] \\
    \hline
    No-ops & [\text{\# Commentary}] \\
    \hline
  \end{tabular}
\end{table}

The only field that represents the progression of a poker state is the ``actions'' field. This field is an array of string action representations, following the action notation described in table \ref{tab:grammar}. It describes how the state mutates throughout the hand as the actors -- the dealer and the players -- act.

The notations in the ``actions'' field must be sorted in the order of the action. The actions may represent a complete history of the hand or a partial history that does not reach the terminal state (the hand being over).

Each action notation aims to concisely describe the actor, the performed action, and arguments if any. The notation may also optionally include a verbal commentary. In addition, a notation may be composed entirely of a standalone commentary or be an empty string. In general, the action notation satisfies the following grammar: \[[Actor\ Action[\ Arguments...]][\ \#\ Commentary]\] where contents that may be omitted are enclosed in square brackets. The words are separated by single or multiple consecutive whitespace characters, and the notation itself may contain leading or trailing whitespaces.

In poker, there are two types of actors: the dealer (also referred to as nature in stochastic games) and the players. The dealer is represented with a single character string ``d'' while $n$'th player is represented with a string ``p$n$'' where $n$ is a 1-indexed player index. The actor information is often omitted in many other game history file format specifications due to redundancy. In most 2-player board games, keeping track of the actor is quite simple. However, in poker, there are often more than 2 actors that rotate around the table who, depending on the situation, are skipped. Even the actor who opens the first betting street (often referred to as pre-flop or third street) differs depending on the forced bet configurations or up cards. It is therefore difficult for human readers, let alone a machine to handle the logic of inferring the actor for each action.

The standing pat and discarding actions are grouped together due to their similarity. Standing pat is a special case of discarding where a player does not discard any hole cards. This notation accepts an optional argument denoting the cards that are being discarded. An omission of this argument denotes that the player in turn is standing pat while the inclusion denotes that the player discarding.

The showdown action is performed at the end of the hand or after all players go all-in. It represents showing of the hole cards to try to win the pot or after an all-in, or mucking of the hole cards. This action accepts an optional argument of cards which, if supplied, must all be known. The omission represents mucking while the inclusion represents showing.

\subsection{Card Notation}

\begin{table}[!ht]
  \caption{All ranks and their single character representations. \label{tab:ranks}}
  \centering
  \begin{tabular}{|c|c|}
    \hline
    Rank & Character \\
    \hline\hline
    Deuce & 2 \\
    \hline
    Trey & 3 \\
    \hline
    Four & 4 \\
    \hline
    Five & 5 \\
    \hline
    Six & 6 \\
    \hline
    Seven & 7 \\
    \hline
    Eight & 8 \\
    \hline
    Nine & 9 \\
    \hline
    Ten & T \\
    \hline
    Jack & J \\
    \hline
    Queen & Q \\
    \hline
    King & K \\
    \hline
    Ace & A \\
    \hline
    Unknown & ? \\
    \hline
  \end{tabular}
\end{table}

\begin{table}[!ht]
  \caption{All suits and their single character representations. \label{tab:suits}}
  \centering
  \begin{tabular}{|c|c|}
    \hline
    Suit & Character \\
    \hline\hline
    Club & c \\
    \hline
    Diamond & d \\
    \hline
    Heart & h \\
    \hline
    Spade & s \\
    \hline
    Unknown & ? \\
    \hline
  \end{tabular}
\end{table}

A card in the PHH format can either be known or unknown. Each known card must be represented with two characters. The first character represents the rank of the card while the second character represents the suit, as tabulated in Tables \ref{tab:ranks} and \ref{tab:suits}. An unknown card must be represented with two question mark characters: ``??''. Multiple cards must be represented by concatenating individual single-card representations without any separators or delimiters. For showdown actions only, if the hole cards being shown have already been specified during the corresponding player's hole card dealing action, the cards can be written as a single dash character: ``-''.

\section{Optional Fields}

\begin{table*}[!ht]
  \caption{The optional fields and their TOML native types. \label{tab:optionalfieldtypes}}
  \centering
  \begin{tabular}{|c|c|c|}
    \hline
    Field & Name & TOML Native Type \\
    \hline\hline
    Annotator full name or mononym & author & String \\
    \hline
    Event name & event & String \\
    \hline
    Event or organizer URL & url & String \\
    \hline
    Venue street-level address & address & String \\
    \hline
    Venue city & city & String \\
    \hline
    Venue region, state, or province & region & String \\
    \hline
    Venue postal code & postal\_code & String \\
    \hline
    Venue country & country & String \\
    \hline
    Timestamp at the start of the hand & time & Local time \\
    \hline
    IANA time zone \cite{ianatzdb} & time\_zone & String \\
    \hline
    Event day & day & Integer \\
    \hline
    Event month & month & Integer \\
    \hline
    Event year & year & Integer \\
    \hline
    Hand number & hand & Integer \\
    \hline
    Tournament level & level & Integer \\
    \hline
    Players' seat numbers & seats & Array of integers \\
    \hline
    The number of seats & seat\_count & Integer \\
    \hline
    Table number & table & Integer \\
    \hline
    Player full names or mononyms & players & Array of strings \\
    \hline
    Final stacks & finishing\_stacks & Array of integers or floats \\
    \hline
    ISO 4127 \cite{iso4127} currency & currency & String \\
    \hline
    Ante uniformity & ante\_trimming\_status & Boolean \\
    \hline
    Allocated time per action & time\_limit & Integer or float \\
    \hline
    Time banks at the beginning of the hand & time\_banks & Array of integers or floats \\
    \hline
  \end{tabular}
\end{table*}

The optional fields may be left unspecified by the annotator. The complete list of optional fields and their TOML native types are tabulated in Table \ref{tab:optionalfieldtypes}.

The description of finishing stacks is helpful as the parser may not be aware of the granularity of the currency the chips are in or the rake and rake-cap applied in the end. On top of this, in a physical setting where chips are used, depending on the denominations, odd chip situations may arise where the player out of position is given the extra odd chip that cannot be broken further. It is, of course, infeasible to describe all the different chip values each player has in a poker hand history format. These inaccuracies are inherent drawbacks of using purely numerical representations to describe the stack values. It is worth noting that the inconsistencies caused by such circumstances only lead to extremely minor ambiguities in the final stack sizes that should not significantly impact the expected value calculations.

\section{User-defined Fields}

Depending on the use case, the required and optional fields may not be sufficient to capture certain desired levels of detail. The specification allows annotators to define custom fields with a single underscore prefix in the field names. This is also in part to ensure that the parser is compatible with PHH files conforming to the future version of the specification that may introduce new fields.

\section{Benchmarking}

\begin{table}[!ht]
  \caption{The conciseness of each hand history format. \label{tab:benchmarks}}
  \centering
  \begin{tabular}{|c|c|c|c|}
    \hline
    Format & \# newlines / hand & \# words / hand & \# bytes / hand \\
    \hline\hline
    PHH & 9.000 & 101.634 & 544.570 \\
    \hline
    PokerStars & 34.359 & 160.519 & 884.711 \\
    \hline
    Pluribus & 1.000 & 1.000 & 125.141 \\
    \hline
  \end{tabular}
\end{table}

Using the average number of newlines, words, and bytes used to represent a hand, we compare the conciseness of the following three formats: the PHH format, the PokerStars hand history format, and the format used in the supplementary of Brown and Sandholm \cite{doi:10.1126/science.aay2400} in Table \ref{tab:benchmarks}. The dataset of hands explored in this analysis is the 10,000 hands played by the poker AI agent Pluribus, all of which were of the no-limit Texas hold 'em variant. The PokerStars hand history version of the dataset was provided by Wang \cite{pluribuspokerstars} and is used for this comparison. All files exclusively use the ASCII character set.

As expected, the Pluribus's original format is the most concise of all three representations. This format is followed by the PHH file format, which is noticeably more concise than the PokerStars format, primarily thanks to its non-verbal nature in the recording method.

Note that the benchmarks of this sort for other hands included in the supplementary is impossible due to the fact that the Pluribus format only supports no-limit Texas hold 'em in specific initial configurations and no documentation is available on recording hands of different variants and configurations in PokerStars notation. In addition, the concepts used in some of the histories like big blind antes are only recently emerging and therefore not supported on PokerStars.

\begin{table}[!ht]
  \caption{The performance of the open-source PHH file format parser. \label{tab:perf}}
  \centering
  \begin{tabular}{|c|c|}
    \hline
    Format & Throughput (hands/s) \\
    \hline\hline
    PHH & 6801.31 \\
    \hline
  \end{tabular}
\end{table}

The performance of our open-source parser implementation in PokerKit \cite{10287546} is tabulated in Table \ref{tab:perf}. The hands in the supplementary were used to calculate the throughput and the parser process was run on a single CPU of Intel\textregistered\ Core\texttrademark\ i5-4690. All the file contents were loaded into memory prior to being parsed. We utilize Python's built-in TOML file reader to parse the file and perform checks to validate the data. On average, it takes 0.147 milliseconds to parse a single hand. Note that no widely established open-source hand history parser for other file formats is available to compare with.

\section{Validation}

We performed multiple experiments in order to validate the correctness of the PHH format, the sample hand histories, and our open-source parser implementation.

First, all the hands included in the supplementary were validated through round-trip testing where the files were converted into snapshots of game states at each action step. These snapshots were then converted back into hand files and compared against the originals. Next, we developed a poker bot evaluation platform with 4 fixed-policy agents biased toward different poker actions and an agent connected to the Slumbot \cite{slumbot} API, the winning heads-up poker agent in the 2012 Annual Computer Poker Competition. Overall, these agents competed against each other for over ten thousand hands. These hands were then saved and the same round-trip testing method was carried out to these files.

Additionally, the parsed games for the Pluribus \cite{doi:10.1126/science.aay2400} hands were checked for consistency and correctness by comparing the final stacks with the payoffs provided in the dataset of Brown and Sandholm \cite{doi:10.1126/science.aay2400}. Similarly, the most of the televised poker hands from our historical selections and the final table of the 2023 WSOP Event \#43: \$50,000 PPC on Day 5 \cite{2023wsopevent43} had overlays of final player stacks in the source video from which we verified that the actions lead to the corresponding stack values. The actions were also checked for violations of the 2023 WSOP Tournament Rules \cite{2023wsoptourneys}.

\section{Discussion}

The PHH format has the potential to serve as a transformational solution to various computational poker applications involving the storage of poker hands.

\subsection{AI Agents}

The use of the PHH format can greatly benefit AI research, especially those that use reinforcement learning and ensemble learning techniques. The comprehensive hand histories allow for detailed analyses and can be used for training purposes. As previously mentioned, previous AI agents were greatly limited in scope due to the initial state assumptions they made about the game. Development of this file format serves as the crucial first step to set up the development of the poker AI agents that overcome these limitations. Furthermore, using a standardized format to share hand histories played by AI agents can also enhance transparency and enable other researchers to easily validate their findings.

\subsection{Online Casinos}

Online casinos that offer poker games can store hands played on their platforms in the PHH file format for archival purposes that can track player history and facilitate easier internal auditing processes. In the industry, logs of the game actions or table data at each snapshot are typically stored. This makes it difficult to carry out statistical analyses or automate certain tasks that may be of interest to these firms. An area of particular interest in online poker is cheating detection. A more highly structured data format for hand histories can facilitate the development of new tools or methods to discern unusual behaviors and identify tell-tale signs of bots. The same data (or maybe another version with the sensitive information like other players' hole cards censored out) can be provided to the players involved in the hand upon request. As an alternative to using persistent files, the developers can map each field into database table columns to store them in a queryable database.

\subsection{Poker Tools}

The PHH format can become a standard for sharing hands between poker players. One interesting avenue of research is in the creation of an open poker hand database akin to various chess databases which would enable poker enthusiasts to reference different hands. The platform can also serve datasets of poker hands for researchers in the field of computational poker for usage in statistical analyses and AI development.

Crowdsourcing and opening of such a database to the public can open up new avenues of poker strategy research. There are largely two schools of poker playing (and developing imperfect information game AI agents): exploitative and unexploitative plays. The first school -- the exploitative approach -- advocates for modeling opponents' behaviors and adjusting one's strategy to take advantage of imbalance in the opponent's range. The second school advocates for adopting a perfectly balanced strategy that cannot be exploited -- Nash Equilibrium. Professional players generally advocate for a mix of both, while academics have mostly focused on the second school of calculating Nash equilibrium strategies \cite{DBLP:conf/aaai/GilpinS05, DBLP:journals/cacm/BowlingBJT17, doi:10.1126/science.aam6960, doi:10.1126/science.aao1733, doi:10.1126/science.aay2400}. Commonly mentioned issues with resolving of Nash equilibrium are that there may exist multiple Nash equilibriums and that it is not well defined in games of more than 2 players due to possibilities of collusion or cooperation among players. A large database of hand histories can provide data to model an average poker player from which exploitative AI agents can be developed with a more well-defined goal of taking maximum advantage of imbalances in commonly employed human poker strategies by diverging from Nash equilibrium.

\section{Conclusion}

This paper introduces the PHH file format, designed to concisely describe poker hand history while allowing ease in human usage and machine parser implementations. Our solution fills a long-standing gap in the existing literature on the computer poker field. The specification builds on top of the TOML specification \cite{toml}, allowing software systems to leverage the rich ecosystem surrounding the TOML file format. By representing different game state information through fields, the hand can be described in a consistent, structured manner. In the supplementary, we provide 10,088 PHH files in 11 different diverse variants to validate the format's flexibility and effectiveness.

The PHH format is suitable to be used in a variety of poker applications in use cases ranging from poker AI development, hand archival, analytical tools, and teaching materials. A promising future direction for research in the field of computational poker is in the development of automated systems that can digitalize televised poker hands into a formats, such as the PHH format, for further analysis and study. This technology would leverage computer vision and machine learning algorithms to recognize chip amounts, player actions, and dealt cards from video footages and/or onboard graphics. Such systems could greatly enhance the efficiency and accuracy of data collection for poker research, eliminating the manual entry of hand histories and reducing human error.

\bibliographystyle{IEEEtran}
\bibliography{main}

\vfill

\end{document}